\documentclass[10pt,twocolumn,letterpaper]{article}

\usepackage{cvpr}
\usepackage{times}
\usepackage{epsfig}
\usepackage{graphicx}
\usepackage{amsmath}

\usepackage{amssymb}
\usepackage{booktabs}
\usepackage{pgfplots}
\usepackage{color}
\usepackage{chngcntr}
\counterwithout*{table}{section}
\usepackage{enumitem}
\setenumerate[1]{itemsep=0pt,partopsep=0pt,parsep=\parskip,topsep=5pt}
\setitemize[1]{itemsep=0pt,partopsep=0pt,parsep=\parskip,topsep=5pt}
\setdescription{itemsep=0pt,partopsep=0pt,parsep=\parskip,topsep=5pt}

\usepackage[pagebackref=true,breaklinks=true,letterpaper=true,colorlinks,bookmarks=false]{hyperref}

\cvprfinalcopy 

\def\cvprPaperID{538} 

\ifcvprfinal\fi
\begin{document}

\title{Cross Domain Knowledge Transfer for Person Re-identification}

\author{Qiqi Xiao\\
\and
Kelei Cao\\
\and
Haonan Chen\\
\and
Fangyue Peng\\
\and
Chi Zhang\\
}

\maketitle

\begin{abstract}
Person Re-Identification (re-id) is a challenging task in computer vision, especially when there are limited training data from multiple camera views. In this paper, we propose a deep learning based person re-identification method by transferring knowledge of mid-level attribute features and high-level classification features.
Building on the idea that identity classification, attribute recognition and re-identification share the same mid-level semantic representations, they can be trained sequentially by fine-tuning one based on another.
In our framework, we train identity classification and attribute recognition tasks from deep Convolutional Neural Network (dCNN) to learn person information. The information can be transferred to the person re-id task and improves its accuracy by a large margin. Furthermore, a Long Short Term Memory(LSTM) based Recurrent Neural Network (RNN) component is extended by a spacial gate. This component is used in the re-id model to pay attention to certain spacial parts in each recurrent unit.
Experimental results show that our method achieves 78.3\% of rank-1 recognition accuracy on the CUHK03 benchmark.
\end{abstract}

\section{Introduction}
\begin{figure}[t]
\includegraphics[width=0.45\textwidth]{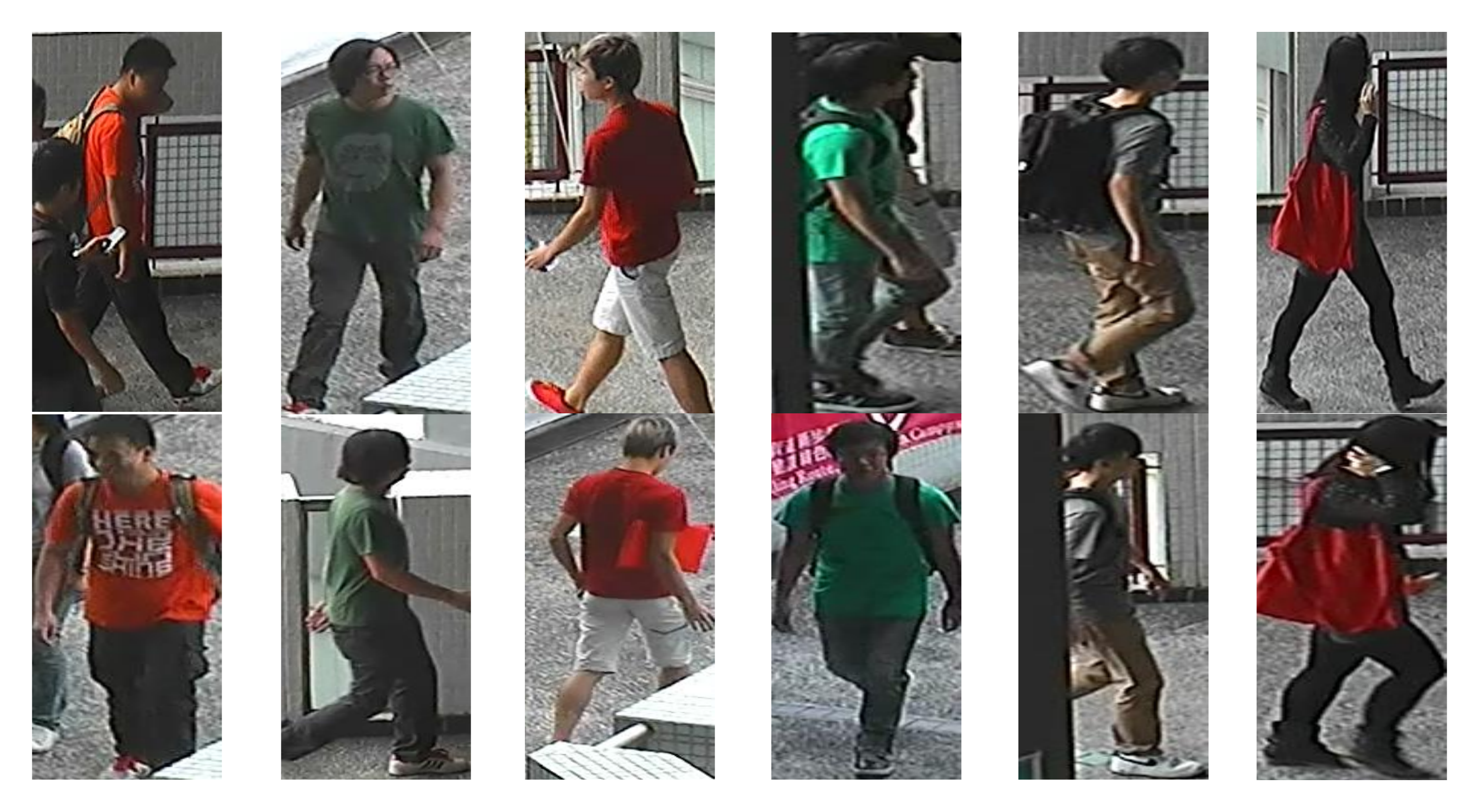}
\caption{Examples of person re-identification dataset, CUHK03. Different identities may be in similar color while the same identity may have different colors and poses due to changes of view points and illumination}
\label{cuhk03_fig}
\end{figure}

Person re-identification (re-id), which aims at matching pedestrians over a set of non-overlapping camera views, is an important and challenging task. 
It provides support for video surveillance, thus saving human labor to search for a certain person and helping improve pedestrian tracking performance. 
However, some issues can make re-identification difficult，such as large variances of individual appearances and poses, environmental changes of illumination and occlusion as well as similarity between different persons, as shown in Figure \ref{cuhk03_fig}.

Some traditional re-id approaches tend to focus on low-level features like colors and shapes to describe the appearance of a person and try to learn a discriminative distance metric to embed the feature space to re-id searching space. These low-level features are not reliable enough because of the above issues. The re-id task can be better treated as a classification task to some extent, which provides with a high-level description of person character. 

The recent development of deep learning has boosted the performance of image classification tasks \cite{He2015Deep,Krizhevsky2012ImageNet,simonyan2014very,szegedy2015going}. A model trained on image classification task is able to learn descriptions of the input image across all levels, which is shown by visualizing technique \cite{zeiler2014visualizing}, and is beneficial for other tasks like object detection \cite{Girshick2015Region} and segmentation \cite{Long2014Fully}. Basically, low-level features are held in bottom layers while mid-level and high-level ones in top layers. These features can be directly used for person re-id tasks and achieves state-of-the-art performance \cite{Liu2016End,Shi2016Embedding,xiao2016learning}.

Previous works have proposed to train the person identity classification task in the first place, and conduct person re-id training to gain more discriminative representation using the same dataset \cite{Liu2016End} or using the subset dataset \cite{xiao2016learning}. In both cases, the classification and re-id tasks can share knowledge.

Meanwhile, human attributes (\eg, having long hair, wearing T-shirt) are mid-level person descriptors that are invariant to changes of the camera angle, view point, illumination or resolution. Attribute recognition has thus long been studied. Its application on person re-id has been explored in recent research works \cite{Layne2012Towards,Su2015Multi,Su2016Deep,Zhao2014Learning} and shows promising results. Therefore, attribute information can also be helpful for person re-identification.

In this paper, we propose a deep learning based knowledge transfer scheme to take advantage of data from cross domain. Each dataset is treated as a domain, which has its own distribution of data. Our experimental results show that the re-id task can benefit from models trained from attribute and classification tasks in different domains.

Furthermore, RNN with LSTM structure is widely used in deep learning based models. Recently, the attention based RNN models have shown excellent performance in areas like phonetic recognition, machine translation, action recognition, \etc. In image related tasks, attention mechanism also achieves the state-of-the-art performance on image captioning \cite{xu2015show} and image generation \cite{gregor2015draw} and has inspired some research works in person re-id \cite{Liu2016End}. We further delve into the principle of attention mask and propose a spacial gate to allow the model to focus on specific part of the image, which is suitable for application scenario of re-id. Quantitative evaluation has been made on re-id benchmarks CUHK03 \cite{Li2014DeepReID} on which we outperforms most state-of-the-art re-id approaches by a large margin.

Moreover, our method has great generalization ability, for the re-id model has a good performance on untrained dataset similar to the real world environment. We also conduct experiments on a larger dataset, showing that the proposed model has the potential for further improvements given more data.

In conclusion, the contribution of this work is three-fold:
\begin{itemize}
    \item First, we propose to train a cross domain knowledge transfer scheme, in which the model is trained on different datasets sequentially that have different data distribution.
    \item Second, we propose a spacial gate based LSTM network for person re-id which helps a lot for our model to have a state-of-the-art performance on the test dataset.
    \item Third, we conduct experiments on various transfer schemes, showing the significance of information transfer. We further conduct experiments with other data and find it possible to further improve the re-id performance using our method.
\end{itemize}

The paper is organized as follows.
Section \ref{relatedwork} reviews related work.
Section \ref{arch} gives detailed explanation of our framework architecture.
Section \ref{experiments} shows our experimental details with results, and gives the comparisons with other methods on several methods on public re-id benchmarks.
Finally, we discuss some interesting phenomenon in Section \ref{discussion}  and we present our conclusion in Section \ref{conclusion}.


\section{Related Work}
\label{relatedwork}
Existing methods on person re-identification generally have two components: feature representation learning and distance metric learning.
We briefly discuss some of these works below.

\paragraph{Deep learning based re-id}
Traditional re-id approaches typically use low-level or mid-level features like colors, shapes and attributes to describe the appearance of a person and learn a good distance metric. Lots of research work falls into this category \cite{Engel2010Person,Hirzer2012Relaxed,koestinger12a,Liao2015Person,Lisanti2014Matching,Paisitkriangkrai2015Learning,Weinberger2009Distance}.
Driven by rapid development of deep learning, deep convolutional neural network (dCNN) is used to extract features from raw images and various methods are proposed to embed dCNN feature to re-id search space, which are regarded as deep metric learning here. The siamese network structure \cite{Ahmed2015An,Li2014DeepReID,mclaughlinrecurrent,Varior2016Gated,Varior2016A,Yi2014Deep} is popular for its incorporation of deep feature extraction and discrimination into a unified framework. Recently, Varior \etal \cite{Varior2016Gated} proposed a subnetwork acting as gate to selectively enhance similar pieces in the whole feature map. Varior \etal \cite{Varior2016A} proposed to divide the image of a person into several rows and feed horizontal clips to an RNN based LSTM, which is followed by the siamese loss. McLaughlin \etal \cite{mclaughlinrecurrent} used the similar LSTM and siamese loss on video-based re-id and achieved the state-of-the-art performance on several video re-id datasets. Siamese approach treats re-id as a classification problem\cite{Zheng2016Person}.

On the other hand, triplet loss is proposed and show great success in learning by ranking problem \cite{Schroff2015FaceNet}. Cheng \etal \cite{cheng2016person} first introduced triplet framework and improve the loss function by adding a predefined margin. Shi \etal \cite{Shi2016Embedding} added a regularization term in the final triplet loss to avoid over-fitting. Liu \etal \cite{Liu2016End} designed a soft attention based LSTM with triplet loss to adapt dCNN feature to re-id task. Wang \etal \cite{wangjoint} combined siamese and triplet architecture into a unified framework. Network with triplet framework achieves state-of-the-art performance in person re-id task.

\begin{figure*}[t]
\includegraphics[width=0.9\textwidth]{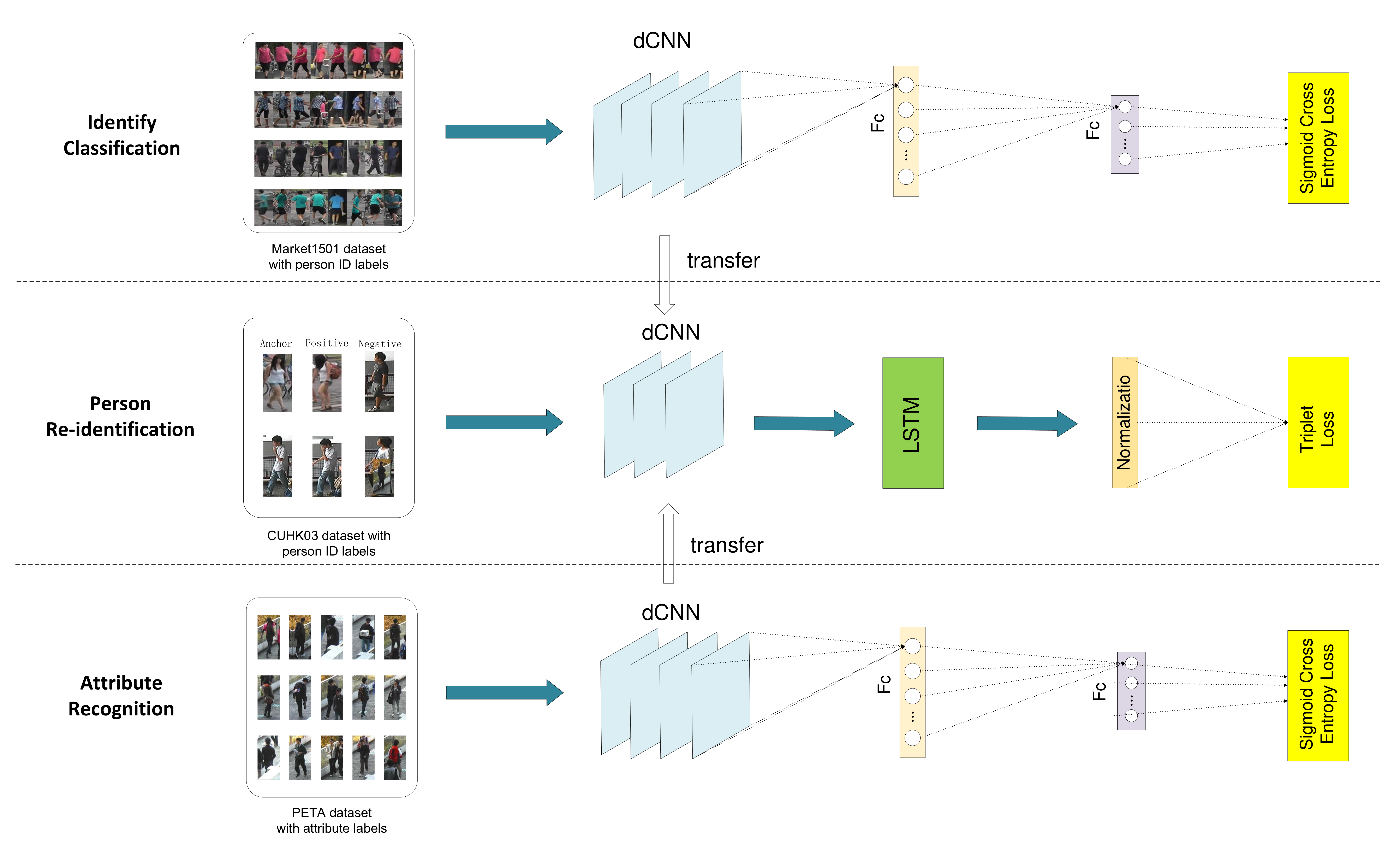}
\caption{Illustration of information transfer framework for person re-identification}
\label{fig:framework}
\end{figure*}

\paragraph{Knowledge transfer for person re-id}
Deep learning based model is extremely data-driven, but large amount of training data demands expensive human effort.
In order to scale to such a situation, transfer learning technique has recently emerged as a promising way to adapt source knowledge to perform new tasks.
For example, network pre-trained on image classification task has shown powerful feature extraction ability and is beneficial for object detection \cite{Girshick2015Region} and segmentation \cite{Long2014Fully}.

Transfer learning technique has been exploited for person re-id in some of the previous works \cite{Avraham2014Learning,Layne2013Domain,Shi2015Transferring}. Specifically, mid-level semantic attribute has long been explored and shown great discriminative power \cite{Layne2012Towards,Su2015Multi,Zhao2014Learning} due to its invariance to viewpoint and illumination changes. Traditional approaches tend to use attribute directly as image descriptor. Recently, Su \etal \cite{Su2016Deep} proposed a three-stage process to learn deep attribute feature from independent attribute dataset \cite{Deng2014Pedestrian} and pedestrian tracking dataset \cite{Lealtaix2015MOTChallenge}, then tested their feature on person re-id datasets \cite{Chen2013Person,Gray2008Viewpoint,Hirzer2011Person}.

\section{Method}
\label{arch}
Networks are trained in three separate domains for different tasks, as illustrated in Figure \ref{fig:framework}, while pretraining task on ImageNet is not shown for its generality. The identity classification task uses Market1501 dataset with person ID labels to perform fully-supervised dCNN training.
The last fully connected layer is set of $1501$ output nodes to classify $1501$ persons. 
In the attribute recognition task, similar structure is used, while the last fully connected layer has $105$ output nodes, taking sigmoid as the loss function. It can recognize whether or not the person has each attribute.
Both the resulting dCNNs produce initial parameters for the person re-id task, providing information with discriminative power to the person re-id task. 
In our re-id architecture, three images are treated as a triplet, in which the second one is a positive sample with the same person id as the first one and the third one is a negative sample with a different person id. The feature extractor consists of a dCNN and an LSTM.
After getting three normalized discriminative features from the feature extractor, the re-id model utilizes triplet loss as the final loss function. Each of those components is elaborated in the following subsections.

\subsection{ResNet for Feature Learning}
\label{resnetsubsection}
In this paper, we adopt ResNet \cite{He2015Deep} structure as the dCNN component for feature learning on the three tasks. ResNet is scalable by stacking the residual component on each other, thus can achieve high demonstration. Most previous works on person re-id use AlexNet \cite{Krizhevsky2012ImageNet} because the re-id dataset is relatively small. On the contrary, as we conduct cross domain transfer learning, a more complex model should be used to fit the large amount of data. We use ResNet-50 to train the person classifier and attribute recognition model. The ResNet-50 network contains 5 stages. The total 5 stages are used for person classification and attribute learning. To transfer the best knowledge to the re-id task, we attempt to use different transfer methods, either holding the 5 stages, only extracting the 4 bottom(near the input) stages or extracting the 3 bottom stages. 
The second one is considered to be most suitable for our task. The first one with 5 stages causes knowledge contained in the top residual component too specific on the training domain, while the third one with 3 stages is too shadow. It can also be supported by our experimental results.

\subsection{Attribute Recognition for Features Enhancing}
In our proposed model, the attribute label of person $p$ is represented as a vector of $k$ binary indicators:\\
\[
A_p = {a_1, a_2, ..., a_K}
\]

where $a_k \in \{0, 1\}, \quad k = 1,2,\dots,K$. Every element of the vector represents whether or not the person $p$ has attribute $k$. A sigmoid cross-entropy loss is used as the final layer, which is calculated as:
\[
\mathcal{L}_{attr} = -\sum_k(a_k\log\hat{a}_k+(1-a_k)\log(1-\hat{a}_k))
\]
The model predicts the probability $\hat{a}_k\in (0,1)$ for attribute $k$.

\subsection{LSTM with the Spacial Gate}
We extract the bottom 4 stages of the pretrained ResNet-50, and complete the feature extractor with a LSTM based RNN component.
Generally speaking, the RNN dynamics can be described using transitions from previous to current hidden states, and the LSTM allows to memorize useful information of several time steps and erase the outdated one.

\begin{figure}[htbp]
\includegraphics[width=0.5\textwidth]{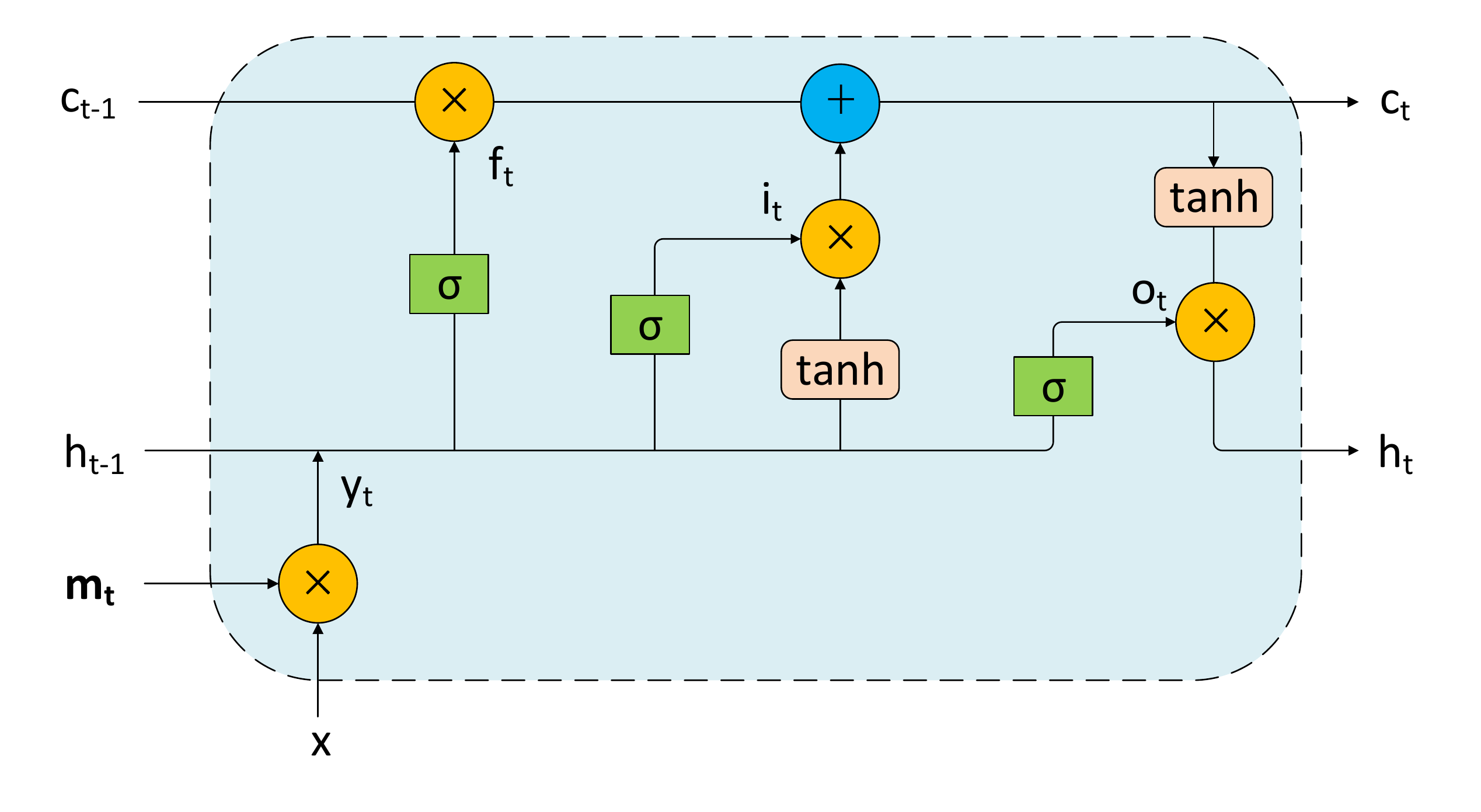}
\caption{Structure of an LSTM unit. We add a mask map s$m_t$ to set a spacial gate and make the unit focus on certain part at each time step.}
\label{fig:lstm}
\end{figure}

The LSTM implementation introduced in \cite{zaremba2014recurrent, xu2015show, sharma2015action, Liu2016End} is used, which is shown in Figure \ref{fig:lstm}. The LSTM unit is conditioned on $c$ channels of the feature map with size $h\times w$ obtained from CNN. The added spacial gate is able to decide which part of the feature map should be used.
The formulations are shown as follows:
\begin{align*}
\left(\begin{array}{cc}
         {i_t}\\
         {f_t}\\
         {o_t}\\
         {g_t}
         \end{array}
\right)
&=
\left(\begin{array}{cc}
         {\sigma}\\
         {\sigma}\\
         {\sigma}\\
         {\tanh}
         \end{array}
\right)
M
\left(\begin{array}{cc}
         {h_{t-1}}\\
         {y_t}
         \end{array}
\right), \\
c_t &= f_t \odot c_{t-1} + i_t \odot g_t, \\
h_t &= o_t \odot \tanh\left(c_t\right), \\
\end{align*}

where $i_t$, $f_t$, $c_t$, $o_t$ and $h_t$ represent input gate, forget gate, cell state, output gate and hidden state respectively. $\sigma$ and $\odot$ are the logistic sigmoid activation and element-wise multiplication respectively. $M$ is an affine transformation with a set of trainable parameters, computing the concatenated result of $h_{t-1}$ and $y_t$. $y_t$ is the result after the input feature map $x$ is multiplied by a normalized mask map $m_t$.
\[
    y_{t,i} = \sum \left(m_t \odot x_i\right), \quad i=1,2,\dots, c,
\]

where $x_i$ is the $i$\emph{th} channel of $x$. The mask map $m_t$ is of size $h\times w$, which should satisfy the constraint $\sum m_t = 1$.

Following \cite{xu2015show}, the initial memory state $c_0$ and hidden state $h_0$ of the LSTM are predicted by an average of feature map per channel fed through two separate multilayer perceptrons, that is:
\begin{align*}
c_0 &= f_c(\frac{1}{h\times w}\sum_{h,w} x), \\
h_0 &= f_h(\frac{1}{h\times w}\sum_{h,w} x)
\end{align*}
This initialization technique is proved to be helpful in our experiments and can make our training process reach the convergent point more easily.

Here we propose four mask maps: global mask, local mask, attention mask and fine-grained attention mask. They are elaborated as follows.

\paragraph{Global mask}
\[
    m_t = \frac{1}{h\times w} J_{h,w}
\]

where $J_{h,w}$ is an all-ones matrix of size $h\times w$. In this situation, the input of the LSTM component $y_t$ keeps the same, \ie, the mean value of $x$ over $h$ and $w$ dimensions. With the global mask, each super-pixel in the feature map contributes equally during the whole process. This type of mask is equivalent to a mean pooling over a channel. 

\paragraph{Local mask}
\begin{align*}
    & m_t(i,j) = \mathbf{1}_A(i), \\
    & A = [\frac{t\cdot h}{n}, \frac{(t+1)\cdot h}{n}), \quad t=0,1,\dots,n,
\end{align*}

where $n$ is the number of time steps and $\mathbf{1}_A$ is the indicator function,
\[
\mathbf{1}_A(x) =
\begin{cases}
1 &\text{if } x \in A, \\
0 &\text{if } x \notin A.
\end{cases}
\]

In this way, one local part of the original feature map is fed into the LSTM component at every time step, thus can extract more discriminative local connections.

Since person structure can be divided into several parts from top to bottom by applying the local mask, the input of the LSTM component is coherently followed one part by another at every time step. Besides, due to pose changes and environment variance across camera views, the key features of a positive pair of person images are not necessarily the same. Meanwhile, it is reasonable to assume a horizontal row-wise correspondence, as all the images we use are resized to a fixed scale and horizontal information has better stability.

Some of the earlier work such as \cite{Varior2016Gated,Varior2016A} have also been inspired by this idea. And compared with the global mask, local mask can better extract person local connections.

\begin{figure}[!h]
\includegraphics[width=0.45\textwidth]{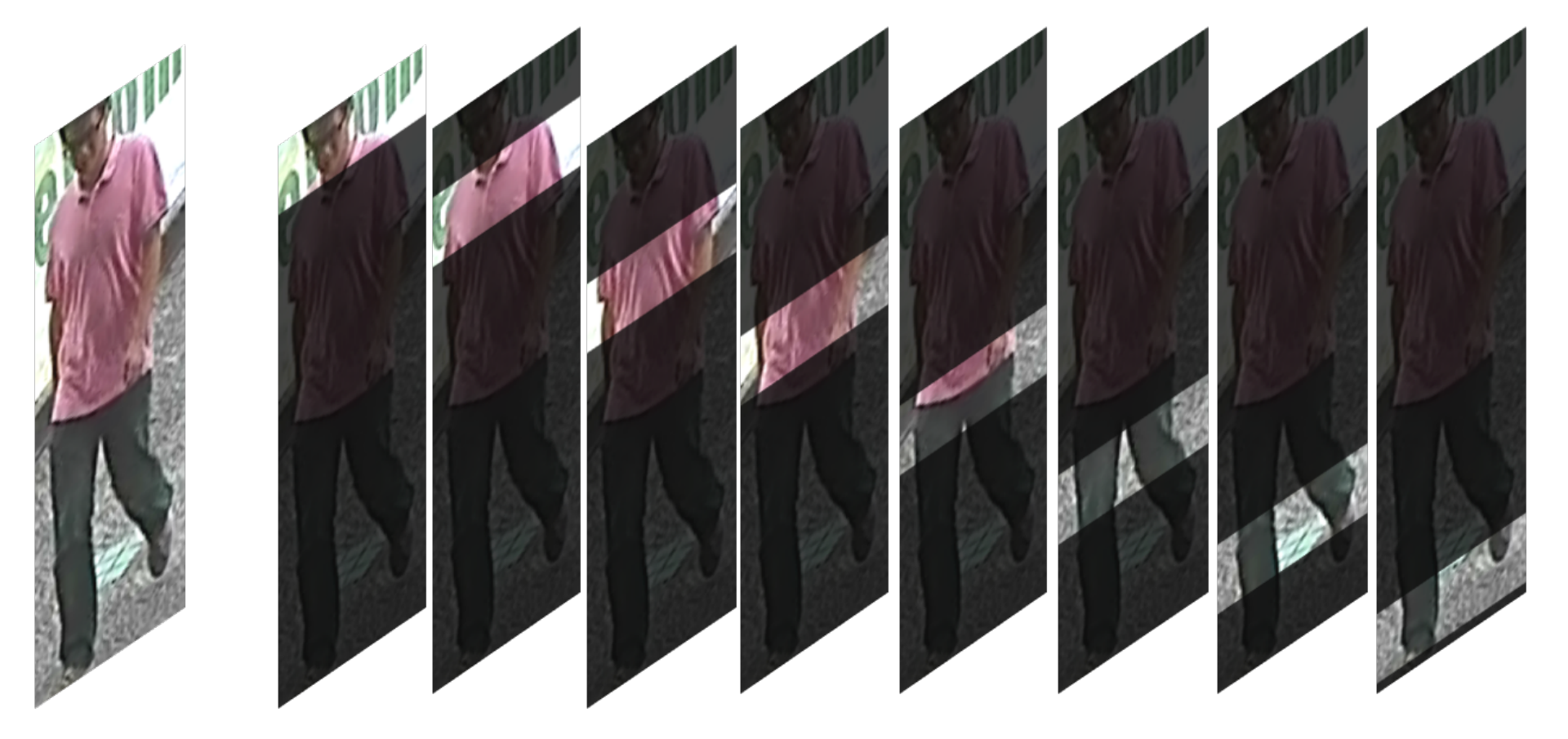}
\caption{Illustration of the local mask map. Note that the masking operation is actually conducted on the feature map.}
\label{fig:maskresult}
\end{figure}

\paragraph{Soft attention mask}
To compute this kind of mask map, the hidden state $h_{t} \in \mathbb{R}^{r}$ ($r$ is the size of hidden state) is repeated on the second and third dimension, resulting in $h_{t}^{(h, w)} \in \mathbb{R}^{r\times h\times w}$. The repeated hidden state of the previous time step $h_{t-1}^{(h, w)}$ is concatenated with the feature map $x \in \mathbb{R}^{c\times h\times w}$. The formula is shown as follows,
\begin{align*}
    m_t &= \text{softmax} \left(N \left(
    \begin{array}{cc}
             {h_{t-1}^{\left(h, w\right)}}\\
             {x}
    \end{array}
    \right)\right),
\end{align*}

where $N$ is a learnable affine matrix.

This learned mask map illustrates that the LSTM can learn to decide which part of the input feature map should be attented, building a comparative attention component. 

\begin{figure}[!h]
\centering
\includegraphics[width=0.4\textwidth]{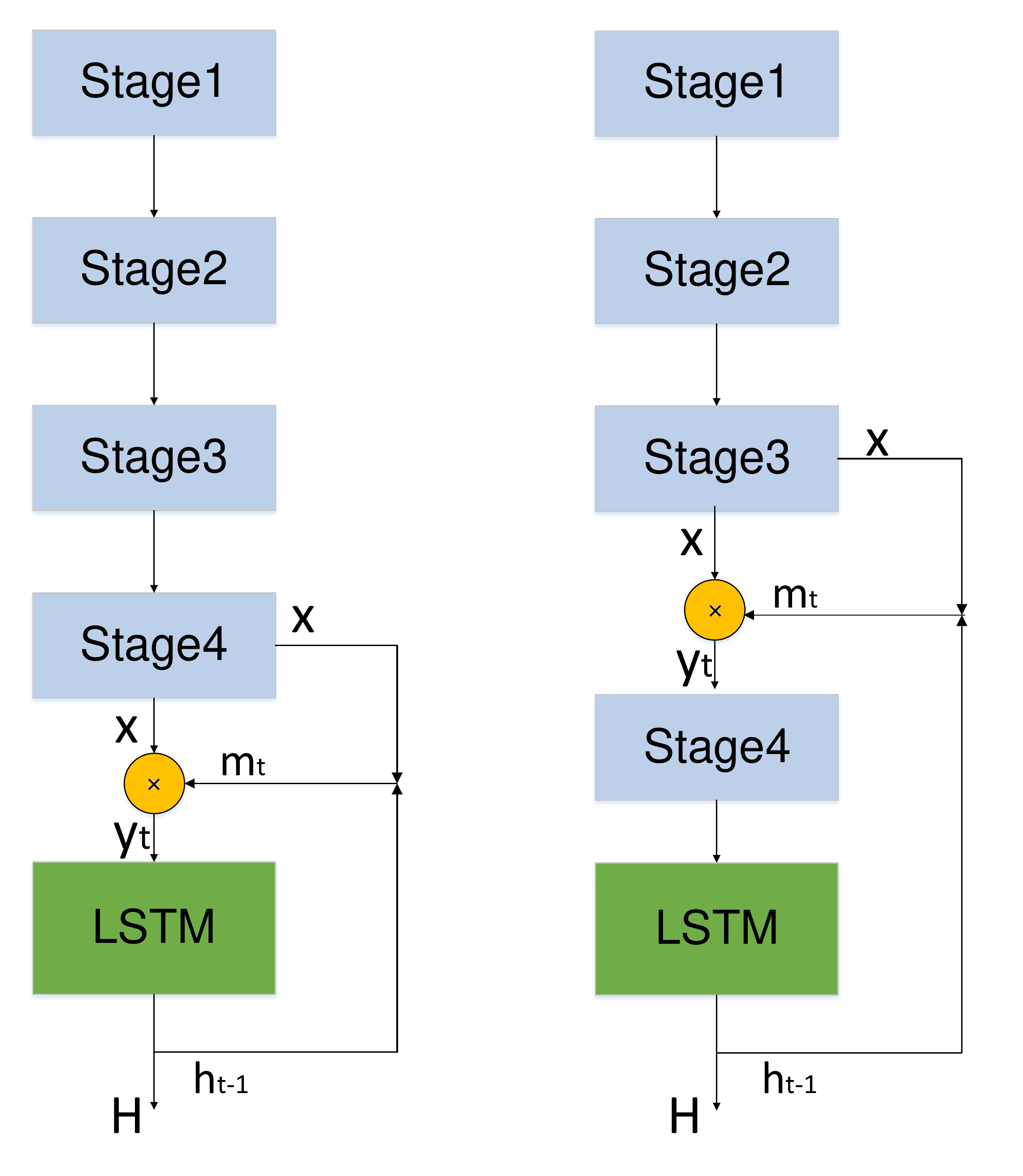}
\caption{Illustration of soft attention mask(left) and fine-grained attention mask(right) in the process}
\label{fig:attention}
\end{figure}

\paragraph{Fine-grained attention mask}
In Figure \ref{fig:attention}, we illustrate our soft attention mask and fine-grained attention mask. Compared with the soft attention mask which adds attention to the feature map obtained from the stage4 of dCNN, fine-grained attention mask means to add attention to the one obtained from the stage3.

The proposal of fine-grained attention mask comes from the deep structure of ResNet, considering of the deepest path. That is to say, the features that deep ResNet extracted are highly abstract and too small in size. Therefore, they can be changed intensively by any attention. 
However, when the feature map is from a more bottom stage, it can hold more spacial information, so attention can be added with a finer grain intensity.

Masked feature map can be sent to the LSTM component after further extracted by layers in stage4 of ResNet-50 at every time step. Further extracting process is practiced because the network of the first three stages does not have strong enough demonstrative ability as mentioned in subsection \ref{resnetsubsection} and complexity reduction needs to be made up.

\subsection{Triplet Selection}
Since our object is to generate features as discriminative as possible, we adopt triplet loss function as our training loss function. In order to compare features, we pick three person images as a training group. In the group, image1 and image2 are of the same person ID and image3 has a different person ID from them. Three images are resized to the same size, and sent to the dCNN and LSTM model separately. In this process, all the three models should share the same weights to ensure the features are extracted in the same way. For a triplet of L2-normalized features $<H, H^+, H^->$, the positive sample $H^+$ is expected to be more similar to $H$ than the negative sample $H^-$, which is formulated that the distance of
$$
\left \| H-H^+\right \|^2 + a < \left \| H-H^-\right \|^2,
$$

where $a$ is the margin we set to indicate the ability the network can distinguish the positive sample and the negative one. As our goal is to improve this ability, the value of $a$ should be able to improve this kind of ability. Out of this goal, the loss function for a triplet in our network is:
\[
\mathcal{L} = \max(d+2a, 0),
\]

where
\[
d = 2\left \| H-H^+\right \|^2 - \left \| H-H^-\right \|^2 - \left \| H^+-H^-\right \|^2.
\]

In the test stage, we apply the trained model to a pair of images. The feature distance between this pair is computed and distances are ranked among all the pairs of queries.

\section{Experiments}
\label{experiments}
We mainly conduct four comparative experiments to test the effectiveness of our model.
\begin{enumerate}
    \item We test three models transferred from different stages of pretrained dCNN.
    \item We conduct experiments fine-tuning from identity classification model and attributes model.
    \item We experiment on the different spacial gates.
    \item We also compare our method with other the state-of-the-art methods on CUHK03 dataset.
\end{enumerate}

\subsection{Datasets}
We use public dataset CUHK03 \cite{Li2014DeepReID} for person re-id, Market1501 \cite{zheng2015scalable} for classification training, PETA \cite{Deng2014Pedestrian} for attribute training.

\paragraph{CUHK03} The dataset contains $14096$ images of $1467$ identities. Each pedestrian is represented from two camera views. Following the evaluation procedure in \cite{Li2014DeepReID}, we take $20$ random splits with $100$ test identities and report the average accuracy. Our experiment is conducted on the hand-labelled dataset.

\paragraph{Market1501} There are more than $25000$ images of $1501$ labeled persons of $6$ camera views in Market1501 dataset. In average, each person contains 17 images with different appearances. Different from CUHK03, a person may be represented from more than $2$ camera views. We use this dataset to train person ID classification model.

\paragraph{PETA} The PEdesTrian Attribute (PETA) dataset is a collection of some small-scale datasets with person images, including some re-id datasets like CUHK. Each image in PETA is labelled with 61 binary attributes and 4 multi-class attributes. We expand the 4 multi-class attributes into 44 binary attributes, following \cite{Su2016Deep}. This results in a binary attribute vector of length 105 for each person. We only use a subset of PETA for attribute training, which is a collection of 3DPeS, CAVIAR4REID, MIT, SARC3D and TownCentre. The dataset varies in camera angles, view points, illumination and resolution.


\subsection{Training Phase Settings}
Each image is resized into $128\times 64$ and then sent into the network. We conduct data augmentation to provide more training data and increase the robustness of trained model. We randomly do horizontal flip, shift, zoom and blur to the original training image in running time. Adam \cite{kingma2014adam} optimizer is used and the initial learning rate is set to $10^{-5}$. The learning rate is increased or decreased according to validation loss. We randomly pick $10\%$ of training data to conduct validation. The margin of triplet loss is set to $a=0.3$ following the setting in \cite{Liu2016End}. And the batch size is set to be $128$.
\subsection{Analysis of the proposed model}

\paragraph{Analysis of different transfer methods}
In this experiment, we first train the model on ImageNet classification and Market-1501 identity classification sequentially. Then we use different transfer methods, to fine-tune with the whole ResNet-50 or only certain stages. An LSTM based RNN in connected and the model is trained on CUHK03 dataset. The average result is shown in Table \ref{tab:transfer}

\begin{table}[htb]
\centering
\begin{tabular}{@{}ccccc@{}}
\toprule
\textbf{Transfer method} & \textbf{Rank1} & \textbf{Rank5} & \textbf{Rank10} & \textbf{Rank20} \\ \midrule
TStage3 & 37.50 & 72.40 & 83.85 & 93.10 \\[0.2cm]
TStage5 & 46.25 & 79.00 & 87.00 & 95.50 \\[0.2cm]
TStage4 & \textbf{72.95} & \textbf{94.30} & \textbf{96.90} & \textbf{98.80} \\[0.2cm]\toprule
\end{tabular}
\caption{Rank1, Rank5, Rank10, and Rank20 recognition rate of different transfer methods on CUHK03 dataset with labeled setting. TStage5 represents transferring the whole ResNet-50 network and TStage3 and TStage4 represent transferring the bottom 3 stages and of 4 stages of ResNet-50.}
\label{tab:transfer}
\end{table}

The result shows that transferring a relatively shallower subnetwork achieves superior result for re-id. This can be explained that the knowledge contained in the top component of dCNN is too specific on the training data domain, thus may be harmful when transferred to other domain with different data distribution. 

\paragraph{Analysis of different domain knowledge}
In this experiment, we analyze the performance of models transferred from different sources. The results are given in Table \ref{tab:results1}.

\begin{table}[htb]
\centering
\begin{tabular}{@{}ccccc@{}}
\toprule
\textbf{Model} & \textbf{Rank1} & \textbf{Rank5} & \textbf{Rank10} & \textbf{Rank20} \\ \midrule
NTransfer & 42.55 & 78.65 & 90.05 & 95.90 \\[0.2cm]
ITransfer & 51.25 & 80.05             & 89.15                 &  94.20                                \\[0.2cm]
ATransfer & 70.25 & 93.95 & \textbf{97.25} & \textbf{98.85} \\[0.2cm]
CTransfer & \textbf{72.95} & \textbf{94.30} & 96.90 & 98.80 \\[0.2cm] \toprule
\end{tabular}
\caption{Rank1, Rank5, Rank10, and Rank20 recognition rate of different levels of information transfer of CUHK03 dataset with labeled setting. The model NTransfer means no information transferred model, and ITransfer, ATransfer, CTransfer mean models transferred from the ImageNet trained, attributes trained and classification trained separately.}
\label{tab:results1}
\end{table}

The baseline model \emph{NTransfer} is trained from scratch, thus contains no information transferred from other domain. It achieves a Rank-1 identification rate of 42.55\%. With the common technique of applying a pre-trained ResNet on ImageNet, we have got an accuracy increase of almost 10\% on \emph{ITransfer}. Both \emph{ATransfer} and \emph{CTransfer} were trained on ResNet-50, and parameters of the bottom four stages were transferred. The mid-level human attribute features, with 105 dimensions as output, could boost the re-id performance to over 70\%, while the high-level classification feature has an even higher performance of 72.95\%. Both models have improved the Rank-5 result to over 93\%. This demonstrates the effectiveness of both classification and attribute information on the re-id task.


\paragraph{Analysis of different spacial gates}
In addition to information transfer, we also conduct experiments to measure the effectiveness of our spacial gates in the LSTM component.
We use time step $n=8$ in this set of experiments, and hidden state size $r=128$. This shape keeps the length-width ratio so that the local relationship can be better reserved. Besides, after passing dCNN, information is extracted and the size of feature map is $8\times 4$ after four pooling layers. The four results are shown in Table \ref{tab:results2}.

\begin{table}[htb]
\centering
\begin{tabular}{@{}ccccc@{}}
\toprule
\textbf{Mask Map} & \textbf{Rank1} & \textbf{Rank5} & \textbf{Rank10} & \textbf{Rank20} \\ \midrule
GM & 72.95          & 94.30          & 96.90          & 98.80          \\[0.2cm]
AM & 73.20          & 94.80          & 98.10          & 99.00          \\[0.2cm]
LM & 74.20          & 94.30          & 97.25          & \textbf{99.20} \\[0.2cm]
FM & \textbf{74.80} & \textbf{94.75} & \textbf{97.80} & 98.95          \\[0.2cm] \toprule
\end{tabular}
\caption{Rank1, Rank5, Rank10, and Rank20 recognition rate of four mask maps in LSTM. GM, AM, LM, FM represents global mask, attention mask, local mask and fine-grained mask respectively.}
\label{tab:results2}
\end{table}

Among the four different mask map types, the global mask simply calculates the mean value of the input feature map, which only achieves $72.95\%$ top-1 score. The local attention only focus on part of the feature map, showing better result compare with both global mask and soft attention mask. Although the soft attention is thought to learn the attention area from training data, it does not improve the re-id recognition accuracy obviously. This may be caused by the complex structure of ResNet. The residual connection of the network has the ability to choose which flow should pass through the network, covering up the function of soft attention to some extent. For this reason, our proposed fine-grained attention mask adds attention to the feature map in a more bottom layer, where the features hold more spacial information. This shows better performance than the original attention model, verifying our assumption.

\subsection{Comparison with state-of-the-art methods}

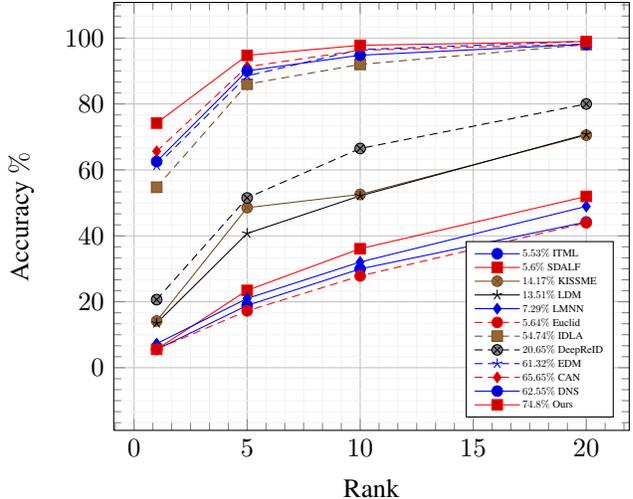
\begin{figure}
\centering
\begin{tikzpicture}
\begin{axis}[
    ymin=-19,
    xlabel={Rank},
    ylabel={Accuracy $\%$},
    grid=both,
    grid style={line width=.1pt, draw=gray!10},
    major grid style={line width=.2pt,draw=gray!50},
    minor tick num=5,
    legend pos=south east,
    legend style={font=\fontsize{10}{5},nodes={scale=0.4}},
    legend cell align=left,
    legend entries={$5.53\%$ ITML,$5.6\%$ SDALF,$14.17\%$ KISSME,$13.51\%$ LDM,$7.29\%$ LMNN,$5.64\%$ Euclid,$54.74\%$ IDLA,$20.65\%$ DeepReID,$61.32\%$ EDM,$65.65\%$ CAN,$62.55\%$ DNS,$74.8\%$ Ours}
  ]
  \addplot table [x=Rank,y=ITML] {cuhk03-sota.txt};
  \addplot table [x=Rank,y=SDALF] {cuhk03-sota.txt};
  \addplot table [x=Rank,y=KISSME] {cuhk03-sota.txt};
  \addplot table [x=Rank,y=LDM] {cuhk03-sota.txt};
  \addplot table [x=Rank,y=LMNN] {cuhk03-sota.txt};
  \addplot table [x=Rank,y=Euclid] {cuhk03-sota.txt};
  \addplot table [x=Rank,y=IDLA] {cuhk03-sota.txt};
  \addplot table [x=Rank,y=DeepReID] {cuhk03-sota.txt};
  \addplot table [x=Rank,y=EDM] {cuhk03-sota.txt};
  \addplot table [x=Rank,y=CAN] {cuhk03-sota.txt};
  \addplot table [x=Rank,y=DNS] {cuhk03-sota.txt};
  \addplot table [x=Rank,y=Ours] {cuhk03-sota.txt};
\end{axis}
\end{tikzpicture}
\caption{Compare with state-of-the-art re-id methods on CUHK03}
\label{fig:cuhksota}
\end{figure}

\paragraph{Performance on CUHK03 dataset}
We compare our method with other state-of-the-art methods on benchmark CUHK03. Traditional methods include ITML \cite{davis2007information}, SDALF \cite{farenzena2010person}, KISSME \cite{koestinger12a}, LDM \cite{guillaumin2009you}, RANK \cite{mcfee2010metric}, LMNN \cite{Weinberger2009Distance} and Euclid \cite{zhao2013unsupervised}. Deep learning methods including IDLA \cite{Ahmed2015An}, DeepReID \cite{Li2014DeepReID}, EDM \cite{Shi2016Embedding}, CAN \cite{Liu2016End}, S-LSTM \cite{Varior2016A}, GSCNN \cite{Varior2016Gated}, DNS \cite{zhang2016learning} and DGD \cite{xiao2016learning}. The results are shown in Figure \ref{fig:cuhksota}. Our proposed model out-performs state-of-the-art re-id methods by a large margin.

\begin{figure}
\centering
\includegraphics[width=0.45\textwidth]{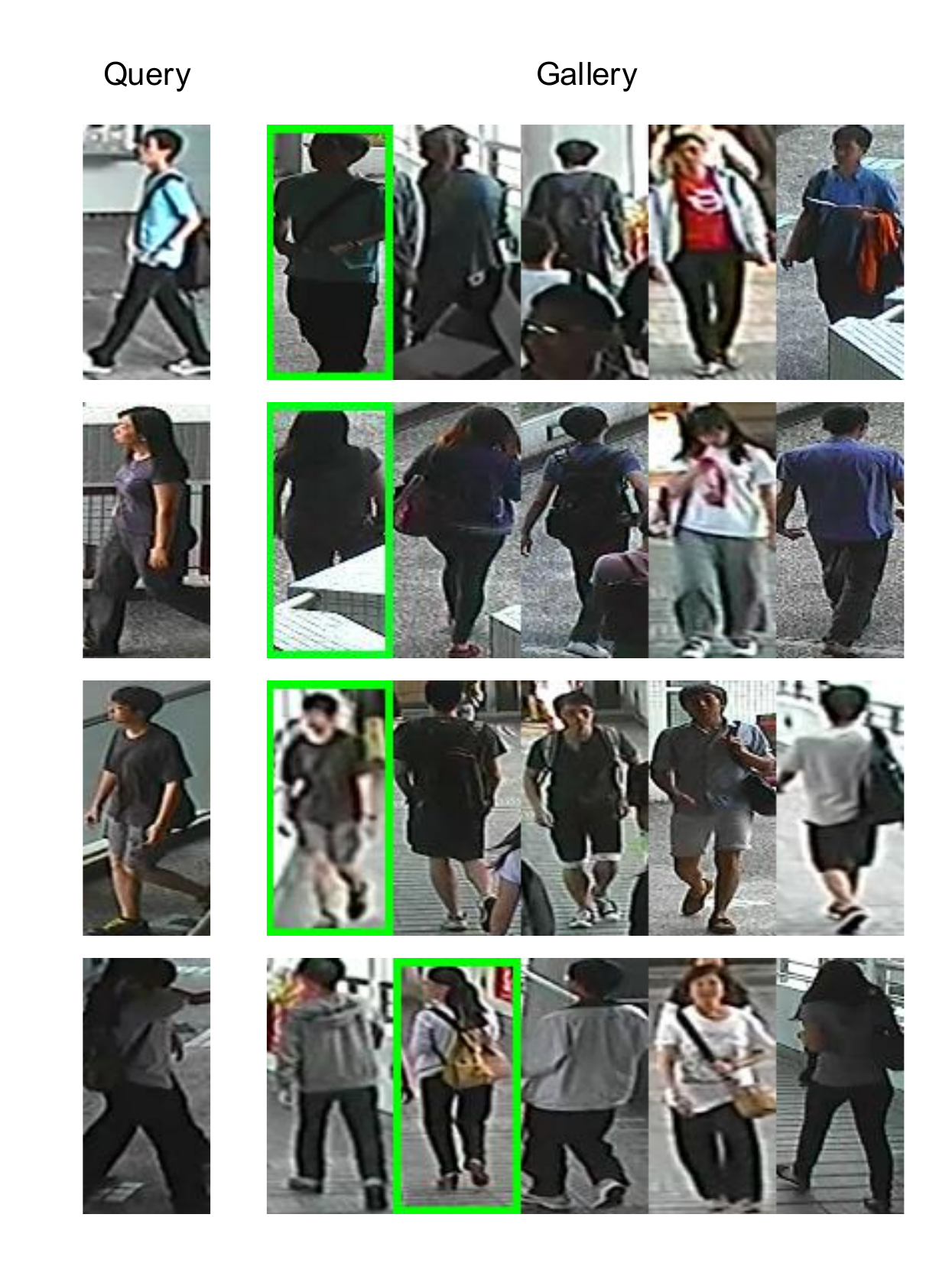}
\caption{The top 5 re-id matches on the CUHK03 test data for a test of 4 queries from the 100 image gallery setting. This shows the local mask based model. Note that the person image in the {\color{green} green} box has the same person id as the query image. Identities of similar appearances are near in re-id search space.}
\label{query_galley}
\end{figure}

\section{Discussion}
\label{discussion}
In this part, we further discuss some interesting aspects that have not been mentioned in the above experimental evaluations.
In the training phase of the classification model,  we use two kinds of datasets. The first one is the Market1501 dataset, which arrives $74.8\%$ of re-identification recognition accuracy. The second one also concludes our own dataset which is crawled on the Internet and annotated by ourselves. The total number of persons reaches above 9000. The accuracy reaches $78.3\%$ under the same condition. Moreover, the recent Domain Guide Dropout re-id model \cite{xiao2016learning} proposed by Tong \etal achieves $75.3\%$ top-1 accuracy on CUHK03 dataset. This method is similar to our classification knowledge transfer while they merge six re-id datasets to train the classification model, containing about 4000 identities. Therefore, more data may take additional accuracy increase. The results are shown in Table \ref{tab:results3}
\begin{table}[htb]
\centering
\begin{tabular}{@{}cccc@{}}
\toprule
Method & Our1500 & JSTL4000 & Our9000 \\ \midrule
Rank1 & 74.80          & 75.30         & \textbf{78.30}            \\[0.2cm] \toprule
\end{tabular}
\caption{Rank1 recognition rate of method using different dataset for classification. Our1500, JSTL4000, Our9000 represent our Market1501 pre-trained method, JSTL method and our 9000 dataset pre-trained method respectively.}
\label{tab:results3}
\end{table}

Some of our example re-id results are shown in Figure \ref{query_galley}. The model is able to learn discriminative feature representations of persons, although some identities are really hard to recognize even for human eyes.

\section{Conclusion}
\label{conclusion}
In this paper, we present an effective information transfer based approach for person re-identification, utilizing identity classification, attribute recognition information to improve the re-id accuracy. Proposed novel spacial gate in LSTM structure takes the advantage of comparative attention to extract intensive person features. Our approach is practical to real-world application such as multiple object tracking, since its excellent performance and serviceability. Experimental results show that our approach greatly improves the performance of person re-identification.

For future work, it is possible to find a better transfer method combining classification and attribute learning to boost re-id performance, or improve the performance of each other simultaneously. Also, person re-id can be integrated with single object tracking to reduce the labor of person recognition. As a real world re-id method, the influence of detection and tracking performance should also be considered in the unified framework.

\clearpage

{\small
\bibliographystyle{ieee}
\bibliography{ref}
}

\end{document}